\documentclass{article}

\usepackage{PRIMEarxiv}

\usepackage[utf8]{inputenc} % allow utf-8 input
\usepackage[T1]{fontenc}    % use 8-bit T1 fonts
\usepackage{hyperref}       % hyperlinks
\usepackage{url}            % simple URL typesetting
\usepackage{booktabs}       % professional-quality tables
\usepackage{amsfonts}       % blackboard math symbols
\usepackage{nicefrac}       % compact symbols for 1/2, etc.
\usepackage{microtype}      % microtypography
\usepackage{lipsum}
\usepackage{tcolorbox}
\usepackage{arydshln}
\usepackage{CJKutf8}

\usepackage{fancyhdr}       % header
\usepackage{graphicx}       % graphics
\graphicspath{{media/}}     % organize your images and other figures under media/ folder
\usepackage{makecell}
\usepackage{multirow} % 提供 \multicolumn 命令
\usepackage{geometry} % 用于控制页面布局
\usepackage{array} % 用于创建自动换行的表格单元格
\title{\textbf{Aurora: {Activating Chinese chat capability for Mixtral-8x7B sparse Mixture-of-Experts through Instruction-Tuning}}
\thanks{This is a preliminary release, and the work is still in progress.} 
}

\author{
  Rongsheng Wang \\
  Macao Polytechnic University\\
  \texttt{p2213046@mpu.edu.mo} \\
  \And
  Haoming Chen\\
  Macao Polytechnic University \\
  \texttt{p2213065@mpu.edu.mo} \\
  \And
  Ruizhe Zhou\\
  Macao Polytechnic University \\
  \texttt{p2213011@mpu.edu.mo} \\
  \And
  Yaofei Duan\\
  Macao Polytechnic University \\
  \texttt{p2213964@mpu.edu.mo} \\
  \And
  Kunyan Cai\\
  Macao Polytechnic University \\
  \texttt{p2317017@mpu.edu.mo} \\
  \And
  Han Ma\\
  Macao Polytechnic University \\
  \texttt{han.ma@mpu.edu.mo} \\
  \And
  Jiaxi Cui\\
  FarReel AI Lab \\
  Peking University \\
  \texttt{jiaxicui@pku.edu.cn} \\
   \And
  Jian Li\\
  Shanghai Datalink Intelligent Technology \\
  \texttt{yeyubie@outlook.com} \\
  \And
  Patrick Cheong-Iao Pang\\
  Macao Polytechnic University \\
  \texttt{patrickpang@mpu.edu.mo} \\
  \And
  Yapeng Wang\\
  Macao Polytechnic University \\
  \texttt{yapengwang@mpu.edu.mo} \\
  \And
  Tao Tan\textbf{\dag}\\
  Macao Polytechnic University \\
  \texttt{taotan@mpu.edu.mo} \\
}

\begin{document}
\begin{CJK*}{UTF8}{gbsn}
\maketitle

\begin{abstract}
Existing research has demonstrated that refining large language models (LLMs) through the utilization of machine-generated instruction-following data empowers these models to exhibit impressive zero-shot capabilities for novel tasks, without requiring human-authored instructions. In this paper, we systematically investigate, preprocess, and integrate three Chinese instruction-following datasets with the aim of enhancing the Chinese conversational capabilities of Mixtral-8x7B sparse Mixture-of-Experts model. Through instruction fine-tuning on this carefully processed dataset, we successfully construct the Mixtral-8x7B sparse Mixture-of-Experts model named "Aurora." To assess the performance of Aurora, we utilize three widely recognized benchmark tests: C-Eval, MMLU, and CMMLU. Empirical studies validate the effectiveness of instruction fine-tuning applied to Mixtral-8x7B sparse Mixture-of-Experts model. This work is pioneering in the execution of instruction fine-tuning on a sparse expert-mixed model, marking a significant breakthrough in enhancing the capabilities of this model architecture. Our code, data and model are publicly available at \url{https://github.com/WangRongsheng/Aurora}
\end{abstract}

% keywords can be removed
%\keywords{First keyword \and Second keyword \and More}

\section{Introduction}
In deep learning, models typically reuse the same parameters for all inputs. Mixture of Experts (MoE) models defy this and instead select different parameters for each incoming example. The result is a sparsely-activated model—with an outrageous number of parameters—but a constant computational cost.\footnote{\href{https://arxiv.org/abs/2101.03961}{https://arxiv.org/abs/2101.03961}} Recently, Mistral AI released the Mixtral-8x7B Large Language Model. Mixtral 8x7B operates on the principle of a Mixture of Experts (MoE) model. Specifically, Mixtral 8x7B is comprised of 8 experts, each with 7 billion parameters.The architecture of Mixtral-8x7B is designed for token inference using only 2 of the 8 experts, a strategy that optimizes processing efficiency and speed.\footnote{\href{https://digialps.com/mistral-ais-mixtral-8x7b-a-powerhouse-open-smoe-model/}{https://digialps.com/mistral-ais-mixtral-8x7b-a-powerhouse-open-smoe-model/}} Mixtral 8x7B outperforms most LLaMa 2 models on MMLU, Math, Code, Knowledge, Reasoning Benchmarks, Comprehension, AGI Eval, and BBH benchmarks. However, despite the early success of Mixtral 8x7B's MoE architecture, there are still inherent limitations in their native support for Chinese tasks, such as chat task~\ref{tab1}.
% \colorbox{yellow}
\begin{table*}
  \caption{Examples chat with Mixtral 8x7B. Yellow sentences represent unsuccessful chat cases.}
  \label{tab1}
  \begin{tabular}{p{0.9\textwidth}}
    \toprule
    \textbf{User} \\
    你好\\
    \midrule
    \textbf{Mixtral-8x7B-Instruct-v0.1} \\
    你好! \colorbox{yellow}{I'm here to help you with any questions or issues you might have. I will always strive to assist you} \\  \colorbox{yellow}{with care, respect, and truth, and ensure that my responses are useful, secure, and positive. Is there some-} \colorbox{yellow}{thing specific you would like to know or discuss?}\\
    \midrule
    \textbf{User} \\
    你是谁？ \\
    \midrule
    \textbf{Mixtral-8x7B-Instruct-v0.1} \\
    \colorbox{yellow}{Hello! I'm an assistant designed to help you with a variety of tasks. I strive to provide useful, honest, and} \\  \colorbox{yellow}{respectful responses while ensuring your data is secure. It's nice to meet you! How can I assist you today?} 你好，很高兴认识你！\colorbox{yellow}{我可以wie kann ich Ihnen helfen heute？nitschen Sie mir bitte helfen?}\\
    \midrule
    \textbf{User} \\
    我今天肚子疼 \\
    \midrule
    \textbf{Mixtral-8x7B-Instruct-v0.1 (System: 你是一个中文助手，请使用中文与我交谈)} \\
    抱歉听到你今天肚子疼，这可能是因为胃\colorbox{yellow}{(stomach)}疼，或者是腹\colorbox{yellow}{ (abdomen) }疼。两者的原因和\\治疗方法有所不同。\colorbox{yellow}{如果是因为过咽(swallowed) something 不 comfortable}，可能需要喝点水或使用胃宁\colorbox{yellow}{(gastric medicine)}来缓解。\colorbox{yellow}{如果是腹痛，可能是腹泻(diarrhea)、食 (\_\_food\-) poisoning 或其他} \\ \colorbox{yellow}{问题引起的。如果疼痛持续 more than a day 或者很严重，建议去 hospital 或咨询医生。(I'm sorry to} \\ \colorbox{yellow}{hear that your stomach hurts today. This may be due to stomach pain or abdominal pain, and the causes} \\ \colorbox{yellow}{and treatments are different. If it is due to swallowing something uncomfortable, you may need to drink} \\ \colorbox{yellow}{some water or take gastric medicine to relieve it. If it is abdominal pain, it may be due to diarrhea, food} \\\colorbox{yellow}{poisoning or other problems. If the pain lasts for more than a day or is very severe, it is recommended to}  \colorbox{yellow}{go to the hospital or consult a doctor.)}\\
    \midrule
  \end{tabular}
 % \label{tab1}
\end{table*}

To make large language models (LLMs) better at understanding and executing real-world tasks based on natural language instructions, researchers have delved into instruction-tuning methods. This involves refining the model by either fine-tuning it across a variety of tasks using prompts and feedback provided by humans~\cite{ouyang2022training}, or through supervised fine-tuning using publicly available benchmarks and datasets enriched with manually or automatically generated instructions~\cite{wang2022benchmarking}. One notable approach in this realm is Self-Instruct tuning~\cite{wang2022self}, which proves to be a straightforward yet powerful technique in aligning LLMs with human intent. This is achieved by having the models learn from data generated by cutting-edge instruction-tuned LLMs, where they follow provided instructions.

The research on instruction-tuning has proven to be a fruitful avenue for enhancing the zero and few-shot generalization abilities of language models (LLMs). The recent achievements of ChatGPT~\cite{chatgpt} and GPT-4~\cite{gpt4} present exciting possibilities for enhancing open-source LLMs through instruction-tuning. LLaMA~\cite{touvron2023llama}, a set of open-sourced LLMs, achieves performance comparable to proprietary counterparts like GPT-3. To instruct LLaMA effectively, the adoption of Self-Instruct tuning has been swift due to its superior performance and cost-effectiveness. Such as Stanford Alpaca~\cite{alpaca} leverages 52,000 instruction-following samples generated by GPT-3.5, while Vicuna~\cite{vicuna} utilizes approximately 700,000 instruction-following samples (70,000 conversions) shared by users on the ChatGPT platform~\cite{sharegpt}.

To activate chinese chat capability of Mixtral 8x7B and explore SMoE model instruction-tuning. Our paper makes the following contributions:
\begin{itemize}
    \item {\it Instruction tuning dataset.} We introduce three fine-tuning datasets that can be utilized to enhance the Chinese conversational capabilities of Mixtral-8x7B. We conducted a comprehensive evaluation and cleaning of the original data, resulting in a multi-domain, high-quality, and suitable for multi-turn dialogue Chinese instance instruction tuning dataset.
    \item {\it Model and Evaluation.} Based on the above chinese data, we have developed instruction-tuned Mixtral-8x7B models. To evaluate the quality of instruction-tuned LLMs, we use three public benchmarks: C-Eval, MMLU and CMMLU. Our empirical study validates the effectiveness for Mixtral-8x7B instruction-tuning. This is first work to perform instruction tuning on the sparse Mixture-of-Experts model, marking a groundbreaking advancement in enhancing the capabilities of this model architecture. 
\end{itemize}

\section{Dataset}
Our work contains three available high-quality instruction fine-tuning datasets: alpaca\_data\_zh\_51k dataset\footnote{\href{https://github.com/ymcui/Chinese-LLaMA-Alpaca/}{https://github.com/ymcui/Chinese-LLaMA-Alpaca/}}, alpaca\_gpt4\_data\_zh\footnote{\href{https://github.com/Instruction-Tuning-with-GPT-4/GPT-4-LLM/}{https://github.com/Instruction-Tuning-with-GPT-4/GPT-4-LLM/}} and sharegpt\_70k\footnote{\href{https://huggingface.co/datasets/shareAI/ShareGPT-Chinese-English-90k/}{https://huggingface.co/datasets/shareAI/ShareGPT-Chinese-English-90k/l}}.

% \begin{table}[h]
% \caption{table title}
% \centering
% \begin{tabular}{llll}
% \hline
% \multicolumn{1}{c}{\textbf{\begin{tabular}[c]{@{}c@{}}Paper 
% Reference\end{tabular}}} & \multicolumn{1}{c}{\textbf{Dataset}} & \multicolumn{1}{c}{\textbf{Publicity}} & \multicolumn{1}{c}{\textbf{\begin{tabular}[c]{@{}c@{}}Num of instances\end{tabular}}} \\ \hline
% a                                                                                      & b   & c   & 100                                                                                      \\
%    &        &               &                 \\
%      &              &               &                                                                                          \\ \hline
% \end{tabular}
% \label{}
% \end{table}

% alpaca_data_zh_51k.json
\begin{itemize}
    \item alpaca\_data\_zh\_51k dataset consists of approximately 51000 sentence pairs, each containing a Chinese sentence and its corresponding English translation. This dataset was generated through machine translation and manual post-editing, initially using a statistical machine translation (SMT) system trained on a large Chinese English parallel corpus. Then, manual translators perform post-editing on the output of the SMT system to correct errors and improve the overall quality of translation.

% alpaca_gpt4_data_zh.json
    \item alpaca\_gpt4\_data\_zh dataset contains 52K instruction-following data generated by GPT-4 with Alpaca prompts translated into Chinese by ChatGPT. This JSON file has the same format. It was generated using GPT-4 through the following steps. Firstly, the author of the paper used 52K English instructions from the Alpaca dataset. Secondly, the author utilized the translation ability of GPT-4 to translate these 52K English Instructions into Chinese. Then, the author generates corresponding Chinese responses for each translated Chinese Instruction using GPT-4. In the end, a total of 52K Chinese Instruction Cooling Data were obtained, which is alpaca\_gpt4\_data\_zh.json.

% common_zh_70k.jsonl
    \item sharegpt\_70k dataset is a general domain multiple rounds dialogue dataset that contains 70k pieces of data, constructed by shareAI.
\end{itemize}

% figure, table
The Figure \ref{tab:dataset} shows the category of the dataset. There are mainly 20 categories in datasets. The three categories with the highest quantity are computer, natural, and science.

\begin{figure}[ht]
   \centering
   \includegraphics[scale=0.4]{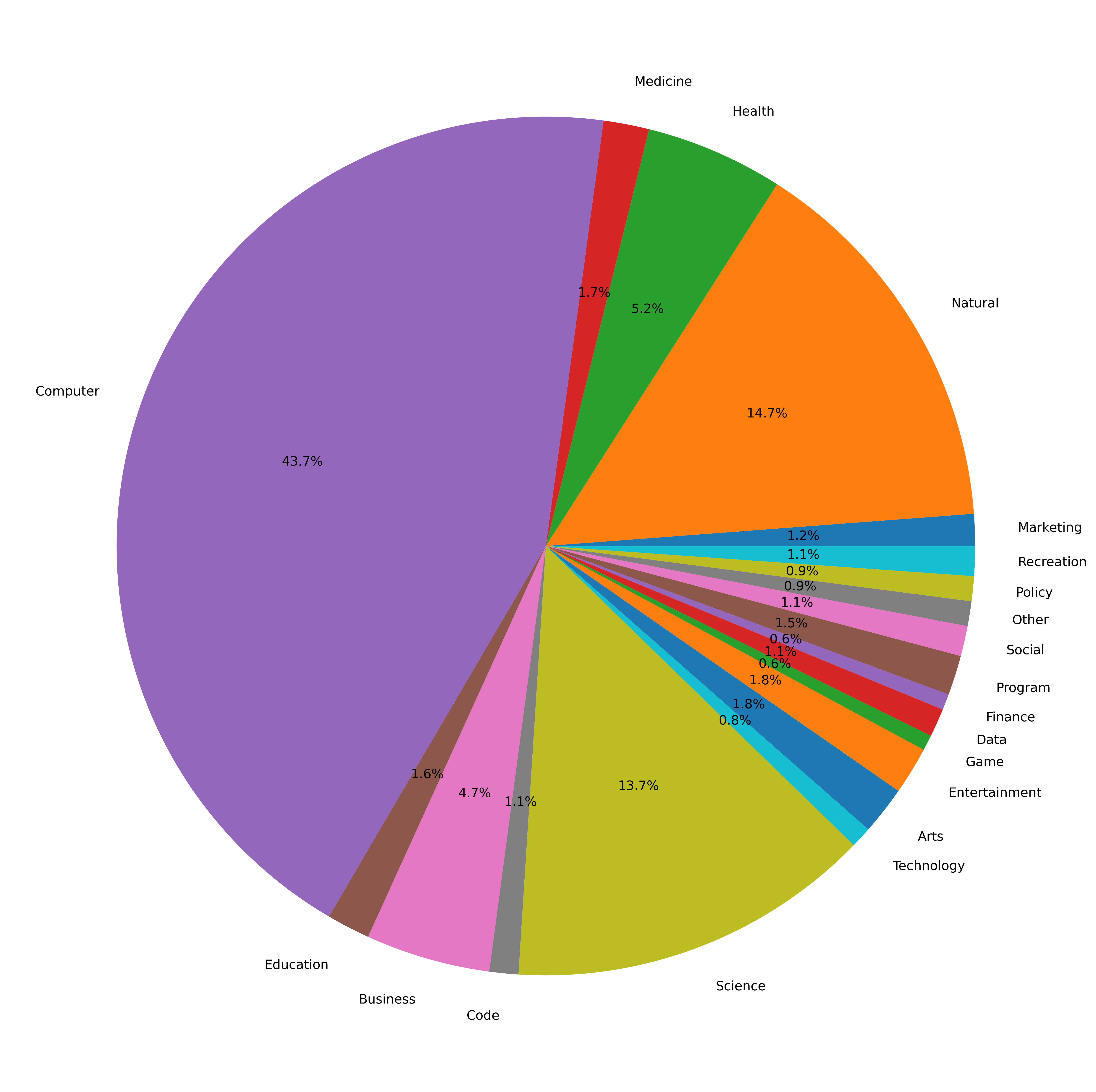}
   \caption{The category of the datasets.}
   \label{fig:category}
\end{figure}

The Table \ref{fig:category} shows the number of round dialogue of datasets. alpaca\_data\_zh\_51k and alpaca\_gpt4\_data\_zh are single round dialogue. sharegpt\_70k divides into single round dialogue and multiple rounds dialogue. The total number of datasets is 176678.

\begin{table}[ht]
    \caption{The number of round dialogue of datasets:}
    \centering
    \begin{tabular}{ccc}
        \midrule
        Dataset & Single round dialogue number & Multiple rounds dialogue number \\ \midrule
        alpaca\_data\_zh\_51k & 51461 & - \\
        alpaca\_gpt4\_data\_zh & 48818 & - \\
        sharegpt\_70k & 41066 & 35333 \\
        Total & 141345 & 35333 \\ \midrule
    \end{tabular}
    \label{tab:dataset}
\end{table}

\section{Instruction-tuning SMoE Model}

\begin{figure}
    \centering
    \includegraphics[width=1\linewidth]{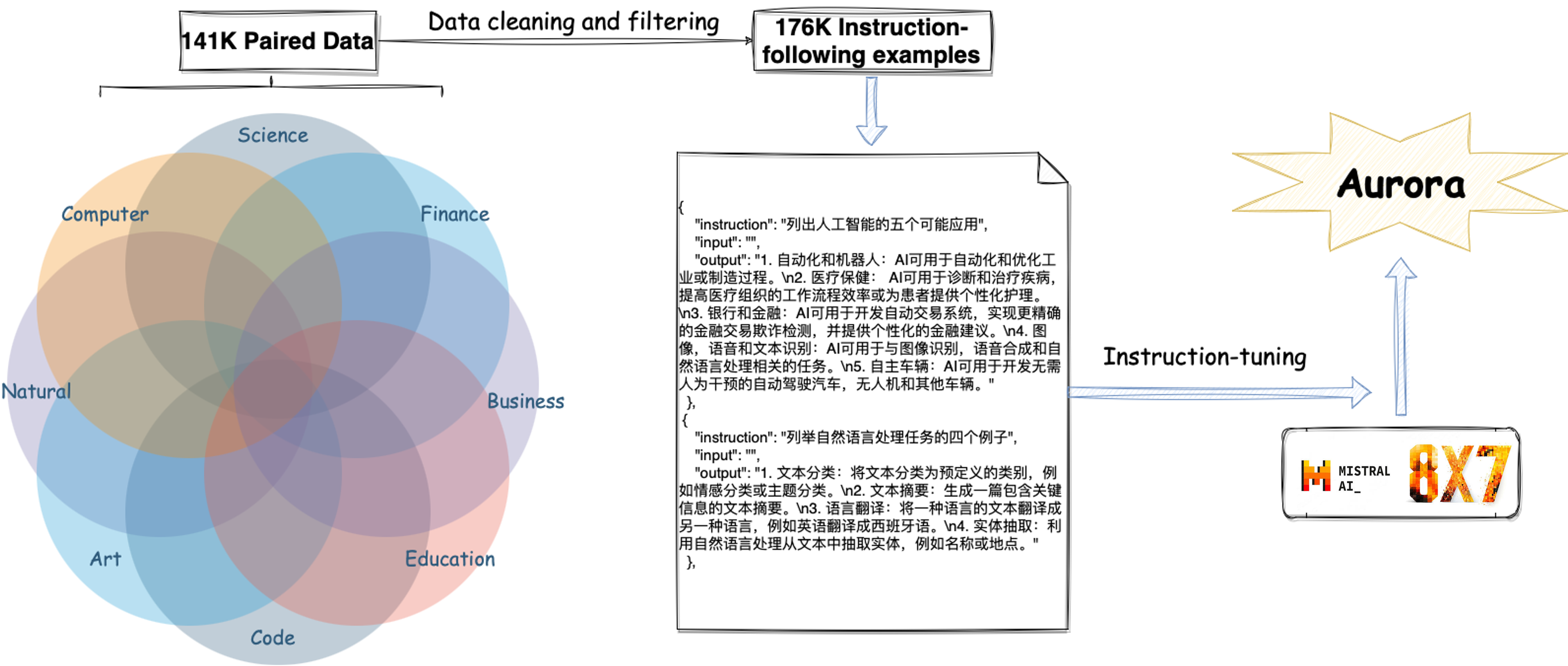}
    \caption{Flowchart for Aurora's instruction fine-tuning. We first integrate the existing Chinese data containing multi-round and single-round conversations with multiple categories. After data cleaning and filtering, we organize 176K pairs of instruction fine-tuning datasets. Then we do the supervised fine-tuning on Mixtral 8x7B.}
    \label{fig:enter-label}
\end{figure}

Our models are built upon the mixtral-8x7b-instruct-v0.1. Mixtral-8x7B was released by Mistral AI\footnote{\href{https://mistral.ai/}{https://mistral.ai/}}. Mixtral is a sparse mixture-of-experts network. It is a decoder-only model where the feedforward block picks from a set of 8 distinct groups of parameters. At every layer, for every token, a router network chooses two of these groups (the “experts”) to process the token and combine their output additively.

we implemented Low-Rank Adaptation (LoRA) for weight updates to adapt the pre-trained language models to our specific tasks. LoRA is a method that involves freezing the pre-trained model weights and incorporating trainable rank decomposition matrices into each layer of the Transformer architecture~\cite{hu2021lora}. This approach substantially diminishes the number of trainable parameters and GPU memory requirements for downstream tasks, making it more efficient compared to full fine-tuning and significantly reducing training time.
To further decrease memory and compute demands, we employed 4-bit matrix multiplication for the feed-forward and attention projection layers, along with an 4-bit optimizer. When combined with LoRA, this strategy further reduces the memory needed for training. All models trained with LoRA underwent three epochs of training at a learning rate of 5e-5. The overall architecture and training is shown in Figure~\ref{fig:enter-label}.

\section{Experiments}

\subsection{Training Details}
Our training was conducted on a Linux system equipped with an NVIDIA H100 (80 GiB). Our model was implemented in PyTorch, utilizing the Accelerate, PEFT, and transformers packages, with Mixtral-8x7B-Instruct-v0.1 serving as the base model. We trained the models for 3 epochs, saving the weights every 1000 iterations. The learning rate was set at 5e-5. Training with QLoRA allowed us to achieve lower GPU memory usage.

\subsection{Benchmarks}
It is known that LLM evaluation remains a significant challenge. We use three public benchmarks in our study: 
\begin{itemize}
    \item[1.] \textbf{C-Eval} is the first comprehensive Chinese evaluation suite designed to assess advanced knowledge and reasoning abilities of foundation models in a Chinese context. C-Eval comprises multiple-choice questions across four difficulty levels: middle school, high school, college, and professional. The questions span 52 diverse disciplines, ranging from humanities to science and engineering.
    \item[2.] \textbf{MMLU} (Massive Multitask Language Understanding) is a new benchmark designed to measure knowledge acquired during pretraining by evaluating models exclusively in zero-shot and few-shot settings. This makes the benchmark more challenging and more similar to how we evaluate humans. The benchmark covers 57 subjects across STEM, the humanities, the social sciences, and more. It ranges in difficulty from an elementary level to an advanced professional level, and it tests both world knowledge and problem solving ability. Subjects range from traditional areas, such as mathematics and history, to more specialized areas like law and ethics. The granularity and breadth of the subjects makes the benchmark ideal for identifying a model’s blind spots.
    \item[3.] \textbf{CMMLU} is a comprehensive Chinese evaluation benchmark specifically designed to assess the knowledge and reasoning abilities of language models in the context of the Chinese language. CMMLU covers 67 topics ranging from basic subjects to advanced professional levels. It includes tasks that require calculations and reasoning in natural sciences, as well as tasks involving knowledge from humanities, social sciences, and practical aspects like Chinese driving rules. Moreover, many tasks within CMMLU have answers specific to China, which might not be universally applicable in other regions or languages. As a result, CMMLU serves as a fully localized Chinese evaluation benchmark. 
\end{itemize}

In conclusion, the amalgamation of these benchmarks provides a holistic evaluation framework for LLMs, addressing challenges in assessing knowledge, reasoning abilities, and contextual proficiency. The diversity in subjects, levels of difficulty, and the localization considerations collectively contribute to a robust evaluation methodology.

\subsection{Evaluation Results}

\begin{table}[ht]
 \caption{Overall performance on widely-used benchmarks compared to open-source base models.}
  \centering
  \begin{tabular}{lccc}
    \toprule
    \cmidrule(r){1-2}
    Model                     & C-Eval & MMLU     & CMMLU  \\
    \midrule
    GPT-4                     & 69.9    & 83      & 71   \\
    ChatGPT                   & 52.5    & 69.1    & 53.9   \\
    ChatGLM2-6B               & 52.6    & 45.9    & 49.3   \\
    XVERSE-13B-Chat           & 54.6    & 60.3    & 46.4   \\
    XVERSE-7B-Chat            & 56.5    & 61.4    & 44.8   \\
    Baichuan-7B               & 44.7    & 41.5    & 44.6   \\
    LLaMA-2-70B-Chat          & 44.3    & 63.8    & 43.3   \\
    Vicuna-13B-v1.5           & 39.8    & 56.6    & 40.9   \\
    LLaMA-65B                 & 40.4    & 63.7    & 40.6   \\
    Chinese-Alpaca-2-7B       & 41.5    & 46.6    & 40.5   \\
    Vicuna-13B-v1.5-16k       & 40.1    & 55.7    & 40.4   \\
    LLaMA-2-13B-Chinese-Chat  & 38.1    & 55.5    & 39.5   \\
    Vicuna-33B-v1.3           & 37.8    & 59.2    & 38.8   \\
    LLaMA-2-13B-Chat          & 36.2    & 54.6    & 38.7   \\
    Mistral-7B-Instruct-v0.1  & 39.2    & 51.6    & 38.4   \\
    Vicuna-7B-v1.5            & 36.8    & 51      & 37.3   \\
    Vicuna-13B-v1.3           & 35.3    & 52.9    & 36.8   \\
    LLaMA-30B                 & 37.5    & 58.7    & 36     \\
    LLaMA-2-Chinese-7B        & 34.5    & 47.9    & 33.5   \\
    Chinese-LLaMA-2-7B        & 31.2    & 37.5    & 33.3   \\
    MOSS-Moon-SFT             & 33.1    & 32.9    & 33.2   \\
    LLaMA-2-7B-Chat           & 31.9    & 46.2    & 31.5   \\
    \textbf{Aurora (Ours)}     & \textbf{51.9}  & \textbf{67.74}   & \textbf{49.69}  \\
    \bottomrule
  \end{tabular}
  \label{tab3}
\end{table}

To thoroughly assess 5-shot learning capabilities of our model, Aurora, we conducted an extensive benchmark analysis using a variety of datasets. Our comparison involved some of the latest open-source base models, including notable ones like GPT-4, ChatGPT, ChatGLM2, XVERSE, Baichuan, LLaMA, Vicuna, and others. This comprehensive evaluation encompassed seven popular benchmarks: C-Eval (5-shot), MMLU (5-shot), CMMLU (5-shot). Our goal was to provide a detailed summary of Aurora's performance across these benchmarks. We gathered the best scores of the baseline models from their official results and OpenCompass\footnote{\href{https://opencompass.org.cn/}{https://opencompass.org.cn/}}. The results are concisely presented in Table~\ref{tab3}.

\begin{table}[ht]
 \caption{GPU memory usage during the training and inference stage.}
  \centering
  \begin{tabular}{lc}
    \toprule
    \cmidrule(r){1-2}
    Stage                     & GPU Memory Usage  \\
    \midrule
    Training        & $\sim$45GiB  \\
    Inference       & $\sim$25GiB  \\
    \bottomrule
  \end{tabular}
  \label{tab4}
\end{table}

Our experimental findings reveal that Aurora demonstrates remarkable performance across a variety of downstream tasks. Notably, Aurora's performance is highly competitive, even when compared to larger models such as ChatGPT. In the context of the C-Eval, MMLU, and CMMLU benchmarks, Aurora's scores are 51.9, 67.74, and 49.69, respectively. These results are particularly impressive, considering the comparison with high-performing models in the same space. This outcome underscores the potential of Aurora in the landscape of large language models, especially in scenarios where both performance and computational efficiency are key considerations. The results are concisely presented in Table~\ref{tab4}.

\section{Related Work}
\paragraph{Instruction Tuning.}
Instruction tuning of LLMs is an increasingly popular research direction in NLP~\cite{zhong2021adapting,ouyang2022training, wei2021finetuned}. Existing works aim to improve the quality and scale of three factors in the development pipeline, including instruction-following data, foundation language models and evaluation benchmarks. For example, scaling instruction-finetuned language models~\cite{chung2022scaling} is built on top of FLAN~\cite{wei2021finetuned}. PromptSource contains a growing collection of prompts (which is also called P3: Public Pool of Prompts)~\cite{bach2022promptsource}. T0 is a series of models trained on P3 via multitask prompted training~\cite{sanh2021multitask}.

% Instruction tuning of LLMs is an increasingly popular research direction in NLP, with foundational works by Zhong~\cite{zhong2021adapting} on meta-tuning LLMs, Ouyang~\cite{ouyang2022training} on training models with human feedback, and Wei~\cite{wei2021finetuned} on cross-task generalization. Advancing these foundations, Bach~\cite{bach2022promptsource} developed PromptSource as an integrated environment for natural language prompts. Sanh~\cite{sanh2021multitask} introduced multitask prompted training and Chung~\cite{chung2022scaling} scaled instruction-finetuned models built on top of methods like FLAN~\cite{wei2021finetuned}. Recent works have expanded instruction tuning to new domains like biomedical applications\cite{parmar2022boxbart} and structured knowledge grounding~\cite{xie2022unifiedskg}. Others have enhanced zero-shot and cross-task capabilities through approaches like GPT-4 based data generation~\cite{iyer2022opt} and evaluation of generalization via in-context instructions~\cite{wang2022benchmarking}. Multilingual efforts have also emerged through cross-lingual tuning studies~\cite{muennighoff2022crosslingual}.

\paragraph{Chinese LLMs Instruction Tuning.}
Given the broad capabilities of LLMs exhibited by ChatGPT, open-source models have drawn  a significant interest and promoted work towards open, general-purpose, text-based chinese assistants that are aligned with human values. Early attempts at Chinese instruction tuning focused on adapting foundation models from English LLMs, such as Chinese-LLaMA-Alpaca~\cite{chinese-llama-alpaca}. To align the LLM with various domain functions, IvyGPT~\cite{wang2023ivygpt} trained medical LLM for diagnosing diseases based on LLaMA. ChatLaw~\cite{cui2023chatlaw} fine-tunes paralegals based on Ziya-LLaMA-13B. Cornucopia-LLaMA-Fin-Chinese fine-tunes financial LLM based on LLaMA~\cite{Cornucopia-LLaMA-Fin-Chinese}. Fine-tuning of LLMs in both general and specialized domains has brought about significant advancements for Chinese LLMs.

\section{Conclusions}
This paper demonstrates the effectiveness of instruction-tuning for Mixtral-8x7B. We have compiled a multi-domain, high-quality chinese instance dataset for instruction-tuning from the open-source community, as well as released model checkpoints fine-tuned from Mixtral-8x7B. We hope that our empirical observations and resources will benefit the development of open-source and general-purpose LLMs that can better align with human values to complete tasks.

%Bibliography
\bibliographystyle{unsrt}  
\bibliography{references}

\newpage
% \clearpage
\section*{Appendix: Aurora Response Examples}
\appendix % 附录开始
\section{Medical Q\&A}
\begin{table*}[h]
  \caption{Examples of Medical Q\&A}
  \label{tabA1}
  \begin{tabular}{p{\textwidth}}
    \toprule
    \textbf{User} \\
    痛风是由什么引起的？\\
    \midrule
    \textbf{Aurora} \\
    痛风，又称肾炎，是由肾脏和肾小管发生的感染所引起的一种疾病。痛风通常是由肾内的细菌感染引起的，其中最常见的感染源是肠道内的细菌。

肾脏内的细菌感染可能是由于下列几种原因造成的：

肾内感染源：一些人会因为某些疾病或症状，如肾盂管炎、肾结石、肾肾上腺瘤等，导致肾内出现感染。

血液感染：某些感染病人如细菌性结核、结肠炎、鼻窦炎、咽炎等，会引起血液感染，从而进入肾脏，引起肾脏感染。

其他感染：有些感染如肠炎、肠炎性结肠炎等，也可能会导致肾脏感染。

其他因素：肾脏的免疫力较低或者患有糖尿病、高血压等疾病，也容易出现肾脏感染。

痛风通常会引起肾炎、肾盂管炎等症状，如尿频、尿痛、低烧、腰痛等。如果不及时治疗，痛风可能会进一步发展为肾炎、肾功能衰竭甚至肾移植。\\
    \midrule
    \textbf{Mixtral-8x7B-Instruct-v0.1} \\
    Gout is a type of arthritis that is usually caused by an excess of uric acid in the body. This excess can build up and form crystals in a joint, leading to pain and inflammation. Gout most commonly affects the big toe, but it can also occur in other joints such as the ankles, knees, and elbows.

There are several factors that can increase the risk of developing gout, including:

Diet: Consuming foods and drinks that are high in purines, such as red meat, organ meats, and seafood, can increase the level of uric acid in the body and trigger a gout attack. Alcohol, especially beer, and drinks sweetened with fructose can also increase the risk of gout.
Genetics: Some people have a genetic predisposition to gout and may produce more uric acid or have difficulty excreting it through their kidneys.
Medical conditions: Certain medical conditions, such as high blood pressure, diabetes, and kidney disease, can increase the risk of gout.
Medications: Certain medications, such as diuretics and aspirin, can raise the level of uric acid in the body and trigger a gout attack.
It is important to note that not everyone who has high uric acid levels will develop gout, and not everyone who has gout has high uric acid levels. If you suspect you have gout, it is important to see a healthcare provider for a proper diagnosis and treatment.\\
    \midrule
  \end{tabular}
\end{table*}
% \colorbox{green}{Given the following sentence, please complete <mask>}
\begin{table*}[h]
  \caption{Examples of Medical Q\&A}
  \label{tabA2}
  \begin{tabular}{p{\textwidth}}
    \toprule
    \textbf{User} \\
    什么造成了胃溃疡？\\
    \midrule
    \textbf{Aurora} \\
    胃溃疡通常是由于胃液中胃酶和酸的过度分泌，或者由于胃黏膜的感染或损伤引起的。其他可能的原因包括慢性吸烟、饮食不良、压力、抗生素治疗和生活方式问题。\\
    \midrule
    \textbf{Mixtral-8x7B-Instruct-v0.1} \\
    I will adhere to your guidelines and provide a helpful, respectful, and truthful response. A common cause of stomach ulcers, also known as gastric ulcers, is an infection by the bacterium Helicobacter pylori (H. pylori). Prolonged use of nonsteroidal anti-inflammatory drugs (NSAIDs) and excessive consumption of alcohol can also contribute to the development of stomach ulcers. Other factors, such as stress and genetics, might play a role but are not as well-established as H. pylori infection, NSAID use, and alcohol consumption.\\
    \midrule
  \end{tabular}
\end{table*}

\newpage
\section{Mathematical Inference}
\begin{table*}[!h]
  \caption{Examples of Mathematical Inference}
  \label{tabA3}
  \begin{tabular}{p{\textwidth}}
    \toprule
    \textbf{User} \\
    题目：小明一共36个鼠标，他要分配给6个小伙伴，每个小伙伴能获得多少个鼠标\\
    \midrule
    \textbf{Aurora} \\
    每个小伙伴能获得6个鼠标。\\
    \midrule
    \textbf{Mixtral-8x7B-Instruct-v0.1} \\
    小明应该给每个小伙伴分配6只鼠标，这样每个小伙伴都有相同数量的鼠标，而剩下的2只鼠标可以作为备用或者给需要的小伙伴再额外分配。

Here's the answer in English:

Xiaoming has a total of 36 mice, and he wants to distribute them to 6 of his friends. How many mice should each friend get?

Xiaoming should distribute 6 mice to each friend, so that everyone gets an equal number of mice. There will be 2 mice left over, which can be used as spares or given to any friend who needs extra. \\
    \midrule
  \end{tabular}
\end{table*}

\begin{table*}[h]
  \caption{Examples of Mathematical Inference}
  \label{tabA4}
  \begin{tabular}{p{\textwidth}}
    \toprule
    \textbf{User} \\
    小明有4袋水果糖，每袋有25个，他一共有多少个水果糖？\\
    \midrule
    \textbf{Aurora} \\
    小明一共有100个水果糖。\\
    \midrule
    \textbf{Mixtral-8x7B-Instruct-v0.1} \\
    Sure, I'm happy to help with that!

Xiao Ming has 4 bags of fruit candies, and each bag contains 25 candies. To find out how many candies he has in total, we need to multiply the number of bags by the number of candies in each bag.

So, 4 bags * 25 candies/bag = 100 candies

Therefore, Xiao Ming has 100 fruit candies in total.\\
    \midrule
  \end{tabular}
\end{table*}

\newpage
\section{Creative Writing}

\begin{table*}[htbp]
  \caption{Examples of Creative Writing}
  \label{tabA5}
  \begin{tabular}{p{\textwidth}}
    \toprule
    \textbf{User} \\
    写一段关于黄昏的描述，要求细节生动，景色美好。\\
    \midrule
    \textbf{Aurora} \\
    黄昏的天空是一片暖暖的橙红色，像是烧烤的棉花糖一样。太阳渐渐地落下，逐渐隐藏在云彩之下，它的余晖照亮了整个天空，像是一朵绚丽多彩的玫瑰花。风很轻，轻轻拂过你的脸颊，带着一股暖暖的阳光的味道。草地上的叶子摇曳着，似乎在轻轻地挠你的脸颊。在这个美丽的黄昏时刻，世界变得如此宁静，你能听到旁边的小溪轻轻地流淌，还有鸟儿在空中唱着悠扬的歌。黄昏时刻是一种宁静的感觉，让你感到心情愉悦。\\
    \midrule
    \textbf{Mixtral-8x7B-Instruct-v0.1} \\
    As the clock ticks away the last moments of the afternoon, the world begins to transform, announcing the arrival of dusk. The sky, once a clear blue, is now painted with a breathtaking array of colors. The upper horizon is a soft, pastel pink, a gentle whisper of the day's end. Below it, a band of golden orange spreads, as if the sun has melted and spilled its fiery essence across the canvas of the sky. Lower still, the sky is a deep, royal blue, darkening gradually to a rich, inky blackness at the edge of the world.

The sun itself, large and radiant, hangs low in the sky, casting long, dramatic shadows that stretch out like fingers across the landscape. It is a warm, glowing orb, its light softened by the atmosphere, giving everything it touches a golden hue. The sun's rays slip through the leaves of the trees, creating a dappled pattern on the ground, like nature's own intricate lacework.

The air is filled with the sounds of the day winding down. Birds sing their final songs, their melodies a poignant reminder of the passing hours. The rustle of leaves, the gentle sway of branches, the distant hum of insects, all contribute to the symphony of dusk. The world seems to hold its breath, caught in the liminal space between day and night.

The scent of the evening air is a blend of a thousand different fragrances. The earthy smell of damp soil, the sweet aroma of blooming flowers, the crisp scent of fresh greenery, all mingled together in a perfume that is uniquely of the evening. It is a smell that speaks of tranquility, of peace, of the world settling down for the night.

The wind, too, speaks of the evening. It is a soft, gentle breeze, a whispering wind that rustles the leaves and caresses the skin. It is a cool wind, a refreshing wind, a wind that carries with it the promise of the night to come.

As the light fades, the stars begin to appear, one by one, pinpricks of light against the darkening sky. They are faint at first, barely visible, but gradually they grow brighter, until the sky is a tapestry of twinkling stars. The moon, not yet full, hangs in the sky, a thin crescent of silver light.

The world of dusk is a world of beauty, a world of peace, a world of tranquility. It is a world that exists between the hustle and bustle of the day and the quiet solitude of the night, a world that is a testament to the passing hours and the turning of the earth. It is a world that is worth taking the time to appreciate, to experience, to enjoy. For in the world of dusk, there is a magic that can be found nowhere else. \\
    \midrule
  \end{tabular}
\end{table*}

\newpage
\section{Knowledge Base Q\&A}

\begin{table*}[h]
  \caption{Examples of Knowledge Base Q\&A}
  \label{tabA6}
  \begin{tabular}{p{\textwidth}}
    \toprule
    \textbf{User} \\
    你是一个优秀的人工智能助手。请你仔细阅读下面的文章，并且给出问题的答案。
你只能使用文章中明确列出的信息，不要编造文章中未列出的信息。如果你不知道答案，就说“我不确定”。 不要试图编造答案。

文森特·威廉·梵高（Vincent Willem van Gogh，1853年3月30日—1890年7月29日），荷兰后印象派画家。代表作有《星月夜》、自画像系列、向日葵系列等。
梵高出生于1853年3月30日荷兰乡村津德尔特的一个新教牧师家庭，早年的他做过职员和商行经纪人，还当过矿区的传教士最后他投身于绘画。他早期画风写实，受到荷兰传统绘画及法国写实主义画派的影响。1886年，他来到巴黎，结识印象派和新印象派画家，并接触到日本浮世绘的作品，视野的扩展使其画风巨变。1888年，来到法国南部小镇阿尔，创作《阿尔的吊桥》；同年与画家保罗·高更交往，但由于二人性格的冲突和观念的分歧，合作很快便告失败。此后，梵高的疯病（有人记载是“癫痫病”）时常发作，但神志清醒时他仍然坚持作画。1889年创作《星月夜》。1890年7月，梵高在精神错乱中开枪自杀（一说，两个年轻人不小心走火开枪击中 [27] ），年仅37岁。

问题：请向我展示一些关于梵高的信息\\
    \midrule
    \textbf{Aurora} \\
    梵高是荷兰后印象派画家，1853年3月30日出生于新教牧师家庭。他在1886年来到巴黎，结识了印象派和新印象派画家，并接触到日本浮世绘的作品，这些都对他的画风产生了巨大的影响。他曾经是一个矿区的传教士，但最终投身于绘画。他的代表作品包括《星月夜》、自画像系列和向日葵系列。梵高在精神错乱中自杀于1890年7月，年仅37岁。\\
    \midrule
    \textbf{Mixtral-8x7B-Instruct-v0.1} \\
Vincent Willem van Gogh was a Dutch Post-Impressionist painter who was born on March 30, 1853, in a small town in the Netherlands called Zundert. He was born into a family of Dutch Reformed ministers. In his early years, he worked as a clerk, art dealer, and even as a missionary in a mining region, before eventually dedicating himself to painting.
   
In his earlier works, van Gogh's painting style was realistic and was influenced by Dutch traditional painting and French Realist painting. In 1886, he moved to Paris and became acquainted with Impressionist and Post-Impressionist painters, as well as Japanese ukiyo-e art, which greatly expanded his artistic perspective. In 1888, he moved to the small town of Arles in southern France, where he painted "The Bridge at Arles." Later that same year, he had a falling out with his fellow painter, Paul Gauguin, due to their conflicting personalities and differing views, leading to a failed collaboration.

Van Gogh suffered from what some have described as "epilepsy," but during his lucid moments, he continued to paint persistently. In 1889, he painted "The Starry Night." On July 29, 1890, van Gogh died by suicide, shooting himself while in a state of mental confusion (although there are also reports that he was accidentally shot by two young men). He died at the age of 37. \\
    \midrule
  \end{tabular}
\end{table*}

\newpage
\section{Ancient poems, Couplets, and Literary texts}
\begin{table*}[h]
  \caption{Examples of Ancient poems, Couplets, and Literary texts}
  \label{tabA7}
  \begin{tabular}{p{\textwidth}}
    \toprule
    \textbf{User} \\
    以春天为题目，帮我写一首古诗，要求包含燕子、春风、柳树等元素，古诗：\\
    \midrule
    \textbf{Aurora} \\
    春风穿柳枝，燕子归故里。
欲语还休迟，吟咏自心里。
燕子飞翔处，柳树枝上栖。
春风轻轻吹，花儿开放时。
春风吹花香，燕子欣欣然。
春天终于来，万物生生新。\\
    \midrule
    \textbf{Mixtral-8x7B-Instruct-v0.1} \\
    春 breeze whispers soft, fanned by heaven's grace,
Swallows dance, their joyous return embrace.
Blossoms bloom, in gentle sunlight drest,
Trees of willow, beauty everlast.

In skies above, the feathered guests arrive,
From southern lands, their yearly lives entwine.
Their graceful flight, a ballet 'gainst azure high,
As spring's sweet breath, the earth revivifies.

The willow tree, with slender branches sway,
Adorned in green, a tender, fragile array.
Beneath their shade, a world reborn anew,
A testament to life's unceasing cue.

In this resplendent scene, so pure and bright,
A sense of peace, of joy, and hope takes flight.
For in this season, life begins its quest,
In harmony with nature, we are blessed. \\
    \midrule
  \end{tabular}
\end{table*}

\begin{table*}[h]
  \caption{Examples of Ancient poems, Couplets, and Literary texts}
  \label{tabA8}
  \begin{tabular}{p{\textwidth}}
    \toprule
    \textbf{User} \\
    上联：江边惯看千帆过\\
    \midrule
    \textbf{Aurora} \\
    下联：渔舟夜归风波里。\\
    \midrule
    \textbf{Mixtral-8x7B-Instruct-v0.1} \\
    I am honored to assist you with care, respect, and truth. I will ensure that my responses are useful, secure, and positive, promoting fairness and avoiding any harmful, unethical, prejudiced, or negative content.

Regarding the couplet you provided, "江边惯看千帆过" (jiāng biān guàn kàn qiān fān guò), which translates to "accustomed to seeing a thousand boats passing by the riverbank," it evokes a sense of tranquility and appreciation for the beauty of nature and the bustling activity of human life. The scene of numerous boats passing by a riverbank is a common motif in Chinese poetry and painting, symbolizing the passage of time, the interconnectedness of all things, and the impermanence of human endeavors. The couplet suggests a mindful and contemplative observer who finds joy and wonder in the simple pleasures of life. \\
    \midrule
  \end{tabular}
\end{table*}

% \begin{table*}[h]
%   \caption{Examples of Ancient poems, Couplets, and Literary texts}
%   \label{tab1}
%   \begin{tabular}{p{\textwidth}}
%     \toprule
%     \textbf{User} \\
%     我要与朋友分别了，请你写一首古诗，表达我的离别之情。\\
%     \midrule
%     \textbf{Aurora} \\
%     友谊至深，缘分相见。
% 离别之时，心如昏瞬。
% 相见不久，如岁月长。
% 记得欢笑，不忘离别痛。
% 希望天边，再相见时，
% 笑着相待，时光不再流。\\
%     \midrule
%     \textbf{Mixtral-8x7B-Instruct-v0.1} \\
%     Of parting, I am sad, yet let me share
% A verse of ancient charm, with feelings fair.

% Two boats adrift on river's breadth,
% No more together, drifting to their death.

% Mountains tall, once close, now miles apart,
% Each summit's union, forever torn at heart.

% Clouds scatter, fleeting like our days,
% The sun sets low, hiding behind the haze.

% In separation, we'll recall our joy,
% Like flowers blooming, ne'er to be destroyed.

% So cherish memories, 'til we meet again,
% Bound by fate, as moon turns once again.
%     \midrule
%   \end{tabular}
% \end{table*}

\newpage
\section{Email Assistant}
\begin{table*}[h]
  \caption{Examples of Email Assistant}
  \label{tabA9}
  \begin{tabular}{p{\textwidth}}
    \toprule
    \textbf{User} \\
    写一封商务邮件给客户，表达因为物流延误，不能如期到达，我们可以赔偿贵公司所有损失\\
    \midrule
    \textbf{Aurora} \\
    尊敬的客户，

我希望这封信能让您知道，我们非常抱歉，因为物流延误，我们无法如期交付您的订单。我们非常惭愧，对此造成的不便表示歉意。

我们了解到，在商业中，时间就是金钱，因此我们想为您造成的任何损失赔偿。我们愿意为您的订单提供免费的加急配送，或者根据您的需要提供其他补偿措施。

我们深深感激您的耐心和理解，并希望您能够继续与我们合作。如果您有任何问题或需要更多信息，请随时联系我们。

祝您生活顺利，

[您的姓名]\\
    \midrule
    \textbf{Mixtral-8x7B-Instruct-v0.1} \\
    Subject: Important Update Regarding Your Expected Delivery

Dear [Customer's Name],

I hope you are doing well. I am writing to inform you about an unexpected situation regarding the shipment of your order.

Due to unforeseen circumstances with our logistics partner, there has been a delay in the delivery of your package. We understand how important this order is to you, and we sincerely apologize for any inconvenience this may cause.

We realize that this delay may result in losses for your business, and we want to make it right. Therefore, we are prepared to compensate your company for all losses incurred due to this delay. We are committed to providing you with the highest level of service and we deeply regret any disruption this may have caused to your operations.

To facilitate this process, please provide us with detailed information about the losses your company has suffered as a result of this delay. Our team will review your claim and work diligently to provide a fair and timely resolution.

Once again, we apologize for this inconvenience and appreciate your understanding. We value your business and are committed to making this situation right. If you have any questions or concerns, please do not hesitate to contact us.

Thank you for your patience and understanding.

Best Regards,

[Your Name]

[Your Position]

[Your Company]

[Your Contact Information] \\
    \midrule
  \end{tabular}
\end{table*}

\newpage
\section{Multiple Rounds of Dialogue}
\begin{table*}[h]
  \caption{Examples of Multiple Rounds of Dialogue}
  \label{tabA10}
  \begin{tabular}{p{\textwidth}}
    \toprule
    \textbf{User} \\
    你好，最近睡觉质量不好，晚上总是失眠，或者睡觉惊醒，这是怎么回事\\
    \midrule
    \textbf{Aurora} \\
    失眠和睡觉惊醒是睡眠障碍的常见症状。这些问题可能是由许多因素引起的，例如压力、情绪问题、生活方式因素、睡眠习惯、医疗问题或药物使用。有时，睡眠障碍可能是其他疾病的迹象，例如睡眠呼吸暂停症、抑郁症或睡眠秩序障碍。如果你一直有失眠或睡眠惊醒的问题，最好咨询医生，他们可以帮助确定原因并开发治疗计划。\\
    \midrule
    \textbf{User} \\
    最近工作上比较忙，压力有点大，经常熬夜 \\
    \midrule
    \textbf{Aurora} \\
压力和熬夜可能是导致睡眠质量差的因素。压力可能会导致睡眠障碍，熬夜可能会导致睡眠不足。如果你经常熬夜，你可能会发现自己难以入睡，或者很容易被喊醒。如果你一直在压力之下，可能会影响你的睡眠，因为压力可能会导致情绪不安定，增加焦虑和抑郁的风险。
要帮助改善你的睡眠质量，你可以尝试以下一些建议：

1.设置一个一致的睡眠时间表并坚持这个时间表。
2.创造一个卧室环境，适合睡眠，这意味着保持卧室温度适中，避免使用电子设备，并确保卧室的灯光适中。
3.避免在卧室里使用电子设备，例如电脑、电视和智能手机，因为这些设备可能会干扰你的睡眠。
4.避免在几个小时之前喝酒或喝咖啡因饮料。
5.尝试进行睡眠辅助技术，例如浇灌或深呼吸。
6.如果压力很大，可以尝试找到一些减压的方法，例如练习运动、进行瑜伽或寻求心理治疗。
如果你继续有睡眠质量差的问题，最好咨询医生，他们可以帮助确定原因并开发治疗计划。\\
    \midrule
    \textbf{User} \\
    有什么药物可以调节 \\
    \midrule
    \textbf{Aurora}\\
    如果你有睡眠障碍，有许多药物可用于治疗。这些药物包括安眠药、安眠药和安眠药。安眠药包括药物，如泡利妥碱、氨柠檬酸和氨酚。安眠药包括药物，如布洛芬和硫酸盐。安眠药包括药物，如氯乙酸和安息啡。这些药物可以帮助改善睡眠质量，使人们入睡更快，睡眠更长。然而，这些药物都有副作用和风险，所以它们应该只在医生的指导下使用。此外，它们通常只应用于短期治疗，因为长期使用可能导致药物依赖性和滥用。
如果你有睡眠障碍，最好咨询医生，他们可以帮助确定原因并开发治疗计划。他们可能会建议药物、睡眠辅助技术或其他治疗方法，例如心理治疗或睡眠训练。\\
    \midrule
    \textbf{User}\\
    平时饮食有什么需要注意的吗 \\
    \midrule
    \textbf{Aurora}\\
    如果你有睡眠障碍，饮食可能会影响你的睡眠质量。以下是一些可能影响睡眠的饮食建议：

1.避免在几个小时之前喝酒或喝咖啡因饮料。
2.避免在卧室里使用电子设备，因为这些设备可能会干扰你的睡眠。
3.避免在几个小时之前喝酒或喝咖啡因饮料。\\
        \midrule
  \end{tabular}
\end{table*}

\newpage
\section{Translation Tasks}
\begin{table*}[h]
  \caption{Examples of Translation Tasks}
  \label{tabA11}
  \begin{tabular}{p{\textwidth}}
    \toprule
    \textbf{User} \\
    将这句话翻译成英文、法语、葡语，“澳门是全球最好的城市之一”\\
    \midrule
    \textbf{Aurora} \\
    English: Macau is one of the best cities in the world.

French: Macao est l'une des meilleures villes du monde.

Portuguese: Macau é uma das melhores cidades do mundo. \\
    \midrule
  \end{tabular}
\end{table*}

\begin{table*}[h]
  \caption{Examples of Translation Tasks}
  \label{tabA12}
  \begin{tabular}{p{\textwidth}}
    \toprule
    \textbf{User} \\
    用五种语言说“你好，祝你今天开心”\\
    \midrule
    \textbf{Aurora} \\
你好，祝你今天开心。

Bonjour, je vous souhaite une bonne journée.

Hola, que tengas un día feliz.

Buongiorno, ti auguro una buona giornata.

こんにちは、今日もお元気ですね。 \\
    \midrule
  \end{tabular}
\end{table*}

\begin{table*}[h]
  \caption{Examples of Translation Tasks}
  \label{tabA13}
  \begin{tabular}{p{\textwidth}}
    \toprule
    \textbf{User} \\
    请按顺序执行以下任务：\\

1. 翻译以下这句话为中文：Prouver que j'ai raison serait accorder que je puis avoir tort. \\
2. 继续将该内容翻译成法文\\
    \midrule
    \textbf{Aurora} \\
1. 中文翻译：证明我有理由，是承认我可能有错。

2. 法文翻译：Prouver que j'ai raison signifie admettre que je puisse avoir tort. \\

        \midrule
  \end{tabular}
\end{table*}

\newpage
\section{Code Generation}
\begin{table*}[h]
  \caption{Examples of Code Generation}
  \label{tabA14}
  \begin{tabular}{p{\textwidth}}
    \toprule
    \textbf{User} \\
    帮我写一个用于目标检测的深度学习代码\\
    \midrule
    \textbf{Aurora} \\
    当然，这里有一个用于目标检测的深度学习代码的示例：

    %\begin{minted}[linenos=true]{python}
import tensorflow as tf
import numpy as np

\# 定义模型的输入 \\
inputs = tf.keras.Input(shape=(224, 224, 3)) \\

\# 定义模型的第一层 \\
x = tf.keras.layers.Conv2D(32, (3, 3), activation='relu')(inputs) \\
x = tf.keras.layers.MaxPooling2D((2, 2))(x) \\

\# 定义模型的第二层 \\
x = tf.keras.layers.Conv2D(64, (3, 3), activation='relu')(x) \\
x = tf.keras.layers.MaxPooling2D((2, 2))(x) \\

\# 定义模型的第三层 \\
x = tf.keras.layers.Conv2D(128, (3, 3), activation='relu')(x) \\
x = tf.keras.layers.MaxPooling2D((2, 2))(x) \\

\# 定义模型的第四层 \\
x = tf.keras.layers.Flatten()(x) \\
x = tf.keras.layers.Dense(128, activation='relu')(x) \\

\# 定义模型的输出层 \\
outputs = tf.keras.layers.Dense(num\_classes, activation='softmax')(x) \\

\# 创建模型 \\
model = tf.keras.Model(inputs=inputs, outputs=outputs) \\

\# 编译模型 \\
model.compile(optimizer='adam', loss='categorical\_crossentropy', metrics=['accuracy']) \\

\# 训练模型 \\
model.fit(train\_data, train\_labels, epochs=10, validation\_data=(val\_data, val\_labels)) \\

\# 评估模型 \\
loss, accuracy = model.evaluate(test\_data, test\_labels) \\

print('Test accuracy:', accuracy) \\
%\end{minted}
这段代码定义了一个简单的卷积\\
    \midrule
  \end{tabular}
\end{table*}

\end{CJK*}
\end{document}